\documentclass[10pt,twocolumn,letterpaper]{article}

\usepackage{cvpr}         

\usepackage{float}
\usepackage{listings}
\usepackage{enumerate}
\usepackage{amsmath}

\usepackage[pagebackref=true,breaklinks=true,colorlinks,bookmarks=false]{hyperref}

\usepackage{soul}
\usepackage{multirow}
\usepackage{amssymb}
\usepackage{amsmath}
\usepackage{wrapfig}
\usepackage{mathtools}
\usepackage[dvipsnames]{xcolor}
\usepackage{booktabs}
\usepackage{bbm}
\usepackage{bigstrut}
\usepackage{xspace}
\usepackage{euscript}

\definecolor{turquoise}{cmyk}{0.65,0,0.1,0.1}
\definecolor{purple}{rgb}{0.65,0,0.65}
\definecolor{darkgreen}{rgb}{0.0, 0.5, 0.0}
\definecolor{darkred}{rgb}{0.5, 0.0, 0.0}
\definecolor{darkblue}{rgb}{0.0, 0.0, 0.5}
\definecolor{blue}{rgb}{0.0, 0.0, 1.0}
\definecolor{magenta}{rgb}{1.0,0,1.0}

\newcommand{\erase}[1]{}

\newcommand{\hide}[1]{{}}


\newcommand{\method}{{\textit{PartCom}}}

\begin{document}

\title{PartCom: Part Composition Learning for 3D Open-Set Recognition}

\author{Tingyu Weng, Jun Xiao,  Haiyong Jiang,\\
School of Artificial Intelligence, University of Chinese Academy of Sciences\\
{\tt\small wengtingyu18@mails.ucas.ac.cn,
    xiaojun@ucas.ac.cn, haiyong.jiang@ucas.ac.cn}
}

\maketitle


\begin{abstract}
    3D recognition is the foundation of 3D deep learning in many emerging fields, such as autonomous driving and robotics.
Existing 3D methods mainly focus on the recognition of a fixed set of known classes and neglect possible unknown classes during testing. 
These unknown classes may cause serious accidents in safety-critical applications, \eg autonomous driving.
In this work, we make a first attempt to address 3D open-set recognition (OSR) so that a classifier can recognize known classes as well as be aware of unknown classes.
We analyze open-set risks in the 3D domain and point out the overconfidence and under-representation problems that make existing methods perform poorly on the 3D OSR task.
To resolve above problems, we propose a novel part prototype-based OSR method named \method. 
We use part prototypes to represent a 3D shape as a part composition, since a part composition can represent the overall structure of a shape and can help distinguish different known classes and unknown ones. 
Then we formulate two constraints on part prototypes to ensure their effectiveness. 
To reduce open-set risks further, we devise a PUFS module to synthesize unknown features as representatives of unknown samples by mixing up part composite features of different classes.
We conduct experiments on three kinds of 3D OSR tasks based on both CAD shape dataset and scan shape dataset. 
Extensive experiments show that our method is powerful in classifying known classes and unknown ones and can attain much better results than SOTA baselines on all 3D OSR tasks. 
The project will be released.
\vspace{-4pt}

    \label{sec:abs}
\end{abstract}
\section{Introduction} \label{sec:intro}

\begin{figure}[t] 
    \subfloat[]{\label{fig1a} 
    \includegraphics[width=0.95\linewidth]{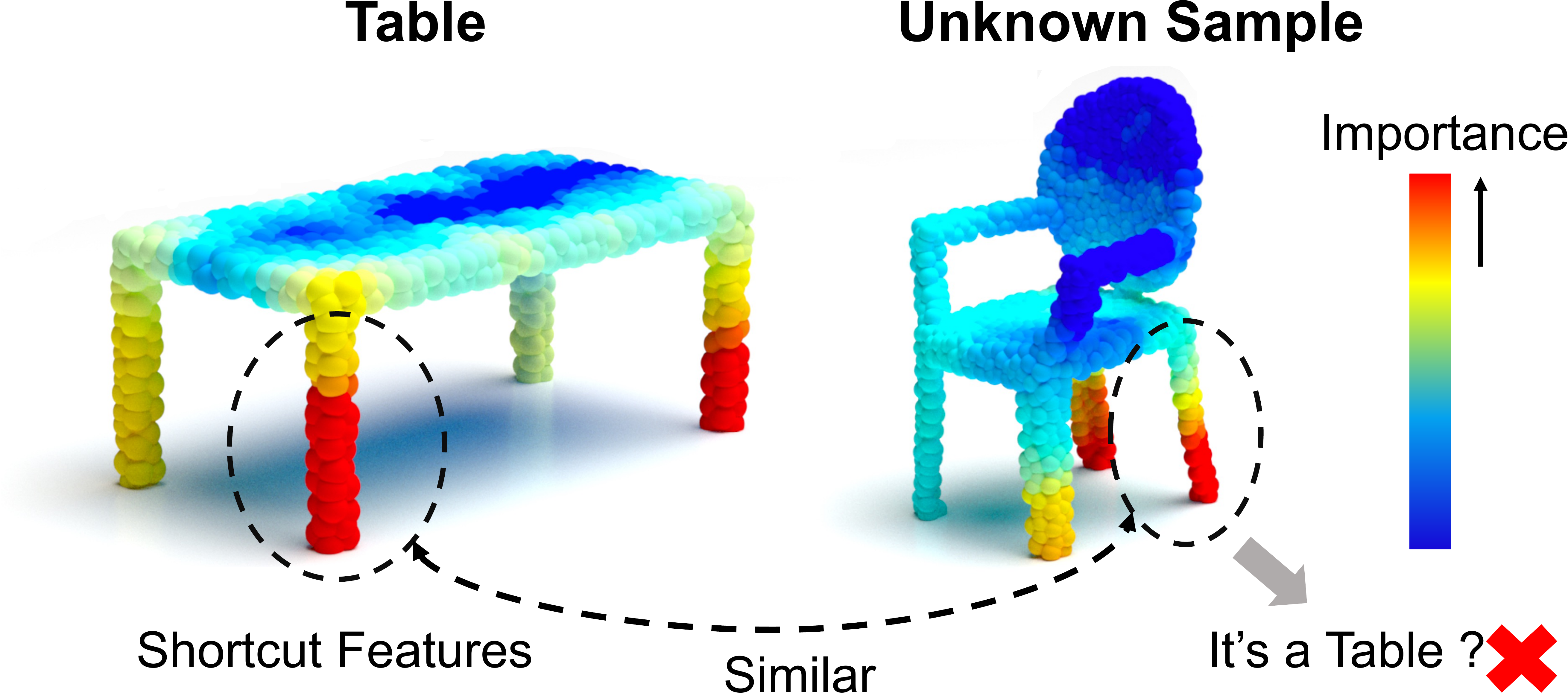}
    }
    \hspace{.4in}
    \subfloat[]{
    \label{fig1b}
    \includegraphics[width=\linewidth]{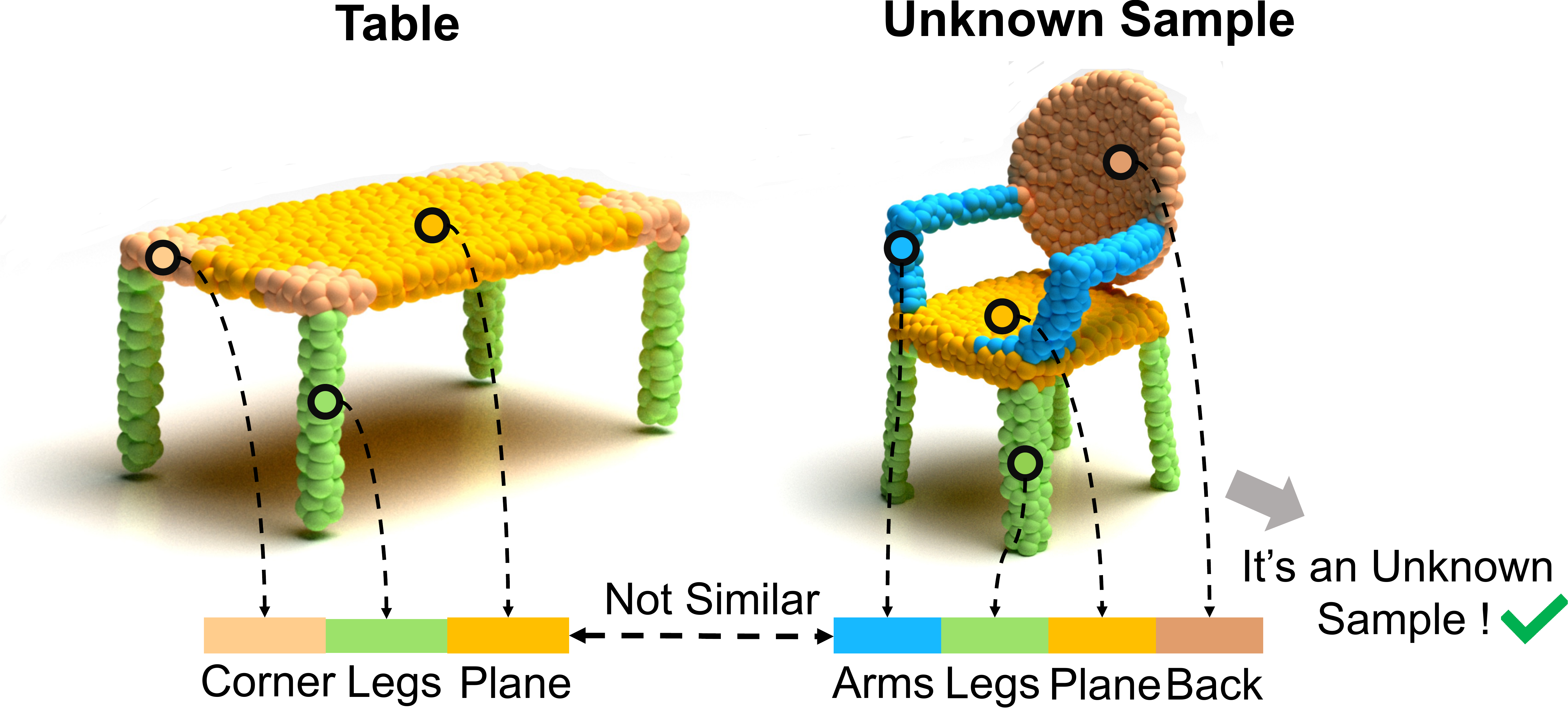}
    }
    \label{fig1}
\caption{(a) The Grad-CAM~\cite{selvaraju2017grad} of a known sample (table) and an unknown sample (chair). The prototype-based methods focus on shortcut features representing local parts and resulting in the incorrect classification of unknown samples. (b) 3D shape can be represented as a part composition. We can easily recognize unknown samples by different part compositions.}
\label{fig1} 
\vspace{-5pt}
\end{figure}

3D deep learning plays an increasingly important role in computer vision fields such as autonomous driving and indoor robotics. 
Recent works~\cite{qi2017pointnet,qi2017pointnet++,zhao2021pointtransformer} have attained a remarkable progress on 3D object recognition.
However, most existing methods assume that the train data and the test data are drawn from the same distribution and share the same set of classes, \ie, the closed-set assumption. 
When encountering a sample from unknown classes, a closed-set 3D classifier usually recognizes the sample as a known class and makes a wrong decision. 
For safety-critical scenarios, \eg, autonomous driving, such errors usually lead to irreparable loss.
We have seen numerous such tragedies~\footnote{https://www.foxnews.com/auto/tesla-smashes-overturned-truck-autopilot,\\ https://www.nytimes.com/2018/03/19/technology/uber-driverless-fatality.html} 
because of unexpected objects on a road and the incorrect recognition of autonomous driving. 
Therefore, it is urging and necessary to investigate the recognition of unknown classes.

The main challenge of extending a 3D classifier to unknown classes is \textbf{the overconfidence problem}~\cite{nguyen2015overconf}, where 3D classifiers tend to recognize unknown sample as a known class with high confidence. 
This is because the learnt feature representation is not distinctive enough for unknown class recognition.
To handle similar problems in 2D image classification, Open Set Recognition (OSR)~\cite{bendale2016towards} is thus proposed to build more robust models capable of recognizing samples of known classes as well as rejecting samples of unknown classes. 
However, an intuitive way of extending the SOTA 2D prototype-based OSR methods~\cite{chen2021ARPL} to the 3D domain faces \textbf{the under-representation issue}. That is to say, learnt prototypes only focus on \textbf{shortcut features}~\cite{hermann2020shapes} representing local regions of a 3D shape, while neglecting important and distinctive information on other regions. 
As visualized in Fig.~\ref{fig1a}, the prototype-based method~\cite{yang2020GCP} determines the class of a shape only by shortcut features, \ie, the legs. 
Therefore it produces the wrong prediction when tested on a shape from an unknown class with legs, \eg, a chair.  

In this work, we propose a part prototype-based 3D OSR method \method~to address the above problems. 
The basic observation is that \textbf{3D shapes from the same class share the common part composition and shape part compositions from different classes differ}. 
For example, a table can be viewed as a composition of legs, planes, and corners.
We can easily distinguish an unknown sample (chair) from a table by comparing their part compositions as shown in Fig.~\ref{fig1b}.
So part composition is unique to each class and can ease the overconfidence of closed-set 3D classifiers.
At the same time, part compositions consider the overall shape structures rather than local parts, thus can alleviate the under-representation of class-specific prototypes.

To this end, our proposed method learns a set of class-specific part prototypes in order to encode a point cloud as part composite feature. 
Then we design two constraints on part prototypes to ensure that they can encode balanced local information 
and be diverse enough. 
Furthermore, we devise a module named PUFS to efficiently synthesize virtual composite features of unknown samples by mixing up part composite features of different classes. 
Classifying virtual composite features as unknown class can encourage a more compact decision boundary of known classes and reduce open-set risks further. 
Our contributions can be summarized as follows.
\begin{itemize}
    \item We analyze open-set risks in the 3D domain and point out two problems that make existing methods perform poorly on 3D OSR task.
    \item We propose a part prototype-based 3D OSR method. To the best of our knowledge, we are the first work to explore 3D open-set recognition.
    \item We dissect potential constraints on part prototypes and develop an optimal transport-based formulation to enforce them. 
    \item We devise a part-based unknown feature synthesis module to reduce open-set risks.
\end{itemize}
We construct various 3D OSR tasks for evaluation. Extensive experiments show that our method significantly outperforms the baseline methods on all 3D OSR tasks.

\begin{figure*}[t] 
    \includegraphics[width=\linewidth]{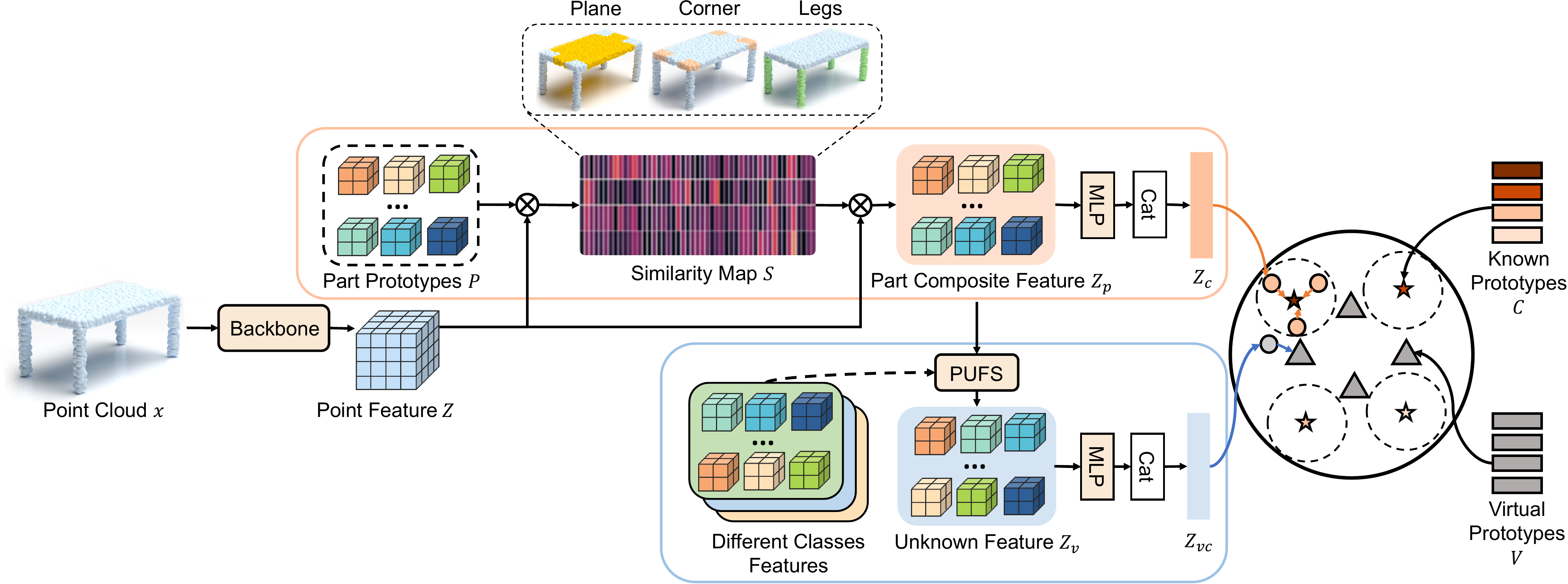}
\caption{\textbf{The overall architecture}. A point cloud is fed into a backbone network to encode point-wise features $Z$. Based on learnable part prototypes $\mathcal{P}$, we construct \textbf{part composite features $Z_P$} as illustrated in the orange box, where point features are aggregated into part composite features according to Eq.~\ref{eq:similarity}. 
We also propose a \textbf{part-based unknown feature synthesis} module to reduce open-set risks as illustrated in the blue box, where part composite features of different known classes are mixed up to synthesize unknown feature of possible unknown classes according to Eq.~\ref{eq:mix}. 
Finally, we learn to classify known features and virtual unknown features by calculating their similarity score with known and virtual unknown prototypes.
} 
\label{fig: pipeline} 
\end{figure*}
\section{Related Work} \label{sec:rw}

\subsection{3D Object Recognition}
Deep learning based methods has transformed the landscape of 3D object recognition and achieve remarkable performance. 
For example, some early works~\cite{wang2017ocnn, choy20194minkowski} develop efficient 3D convolution operations for 3D representation learning inspired by the success of 2D convolutions.
Another line of pioneering works focus on representation learning on point clouds, such as PointNet~\cite{qi2017pointnet} and PointNet++~\cite{qi2017pointnet++}. 
Later methods improve the performance by proposing more flexible network layers~\cite{li2018pointcnn,wang2019dgcnn,zhao2021pointtransformer}.
All these methods assume the training dataset and the testing dataset share a common set of labels, \emph{i.e.}, the closed-set assumption. 
However, this assumption does not hold in most times in real world applications and it is easy to encounter unknown objects during testing, \emph{e.g.,} unknown animals on a road when driving. 
In this work, we make a first attempt to the open-set recognition of the 3D domain. 

\subsection{2D Open-set Recognition}
Deep learning based 2D open-set recognition can be divided into three categories, \emph{i.e.,} calibration-based methods, prototype-based methods, and generation-based methods. 
Calibration-based methods focus on calibrating the output logits of each category to avoid overconfidence of closed-set classifiers~\cite{nguyen2015overconf}. For example, Ben et al.~\cite{bendale2016towards} replace the softmax function with OpenMax and calibrate the output logits with Weibull distributions. Zhou et al.~\cite{zhou2021placeholder} calibrate the output probability by learning dummy classifier placeholders located between known classes.
Prototype-based methods~\cite{yang2020GCP} utilize class-specific prototypes to learn more discriminative feature representation and achieve the state-of-the-art results. For example, Chen et al.~\cite{chen2020RPL, chen2021ARPL} use reciprocal points to construct the open-set space and classify a sample as unknown if it is close to reciprocal points.
These methods can cause the under-representation of features because all the data of a category is represented by a single prototype. 
Generation-based methods take another scheme by synthesizing unknown examples with generative neural networks to manage open-set risks. For example, Kong et al.~\cite{kong2021opengan} explore the training data and crowd-sourced data to train a conditional GAN for unknown synthesis.   
To reduce the dependency on external data, some works~\cite{ge2017G-openmax,neal2018counterfactual} instead train a GAN on closed-set images to generate counterfactual images as unknowns.
Most of this line of works require to train additional generative networks. 
Moreover, generation-based methods only work well on simple MNIST images and have difficulty in modeling diverse patterns in natural images, \emph{e.g.,} ImageNet images. 
Note that all of these methods focus on 2D open set problems, whereas our method investigates 3D open-set recognition problems, where the data is more unstructured and the network design is more challenging. 
In this work, we propose to explore consistent part compositions of different 3D shapes to address the under-representation problem of prototype-based methods and mix up different composite features to efficiently generate unknown samples which avoids the drawbacks of generation-based methods.
\section{3D Open-Set Recognition}
In this section, we formalize the problem of \emph{3D open set recognition}. 
Existing 3D recognition algorithms~\cite{qi2017pointnet,qi2017pointnet++} mainly focus on 3D classification in a \emph{closed-set} configuration, where semantic labels of the training set and the testing set come from the same set of classes and follow the same distribution. 
Consider the training set $D_{train}=\{(x_i,y_i)\}_{i=1}^N$ and the testing set $D_{test\_closed}=\{(x_i,y_i)\}_{i=1}^M$, where $x_i$ denotes an input sample and $y_i \in \mathcal{Y_K} = \{1,2,...,K\}$ is the associated label for a sample. 
Here, $x_i$ can be any 3D representation, \eg, 3D point cloud or 3D mesh, and we use point cloud to describe the method in the following texts.  
However, the closed-set assumption does not hold in most real world scenarios, where 3D shapes with unknown labels are everywhere.
For example, unknown objects on the road may lead to a serious car accident.

In order to handle the above problem, we propose to extend the open-set recognition to 3D domain. 
Given the training set $D_{train}$ and the testing set $D_{test\_open}=\{(x_i,y_i)\}_{i=1}^M$, 
an open-set classifier aims to classify 3D samples of known classes $\mathcal{Y_K}$ as well as categorize those unknown samples as unknown class $K+1$,
where $y_i \in (\mathcal{Y_K}\cup\mathcal{Y_U})$ and $\mathcal{Y_U}$ denotes unknown classes that do not appear in the training set. 
Note that class $K+1$ represents all unknown classes $\mathcal{Y_U}$, which can contain more than one class.

\section{Proposed Method} \label{sec:method}
\emph{3D open-set recognition} is challenging in several aspects. 
First, existing 3D classification is based on the closed-set assumption and tends to classify any sample as a known class with high confidence, making it difficult to handle OSR problems. 
Second, an intuitive extension of the SOTA 2D OSR method, \eg, \cite{yang2020GCP}, is confronted with feature shortcuts and only pays attention to specific local parts, thus leading to the under-representation of a 3D shape. 


In this work, we propose a part prototype-based 3D OSR method dubbed \method, which uses the part composition of a 3D shape to address the above problems. 
As shape part compositions have better discrimination among different-class shapes and usually consider the overall shape structures rather than only local parts, it can alleviate the overconfidence problem in closed-set methods and the under-representation problem of prototype-based methods as shown in Fig.~\ref{fig1}.

The overall framework is shown in Fig~\ref{fig: pipeline}. First, we encode point-wise feature of a point cloud with an off-the-shelf backbone (\ie, PointNet++\cite{qi2017pointnet++}). 
Then we aggregate point cloud feature into part composition feature (see Sec.~\ref{subsec:part-proto}) and develop two constraints to ensure the effectiveness of part prototypes (see Sec.~\ref{subsec:part-constr}). Finally, we devise a part-based unknown feature synthesis module to reduce open-set risks further (see Sec.~\ref{subsec:part-virtual}).

\subsection{Part Prototype-based 3D OSR}\label{subsec:part-proto}

3D shapes from the same class have a common part composition, while part compositions of different class shapes vary. We can easily distinguish an unknown sample from known classes by comparing their part compositions as illustrated in Fig.~\ref{fig1b}.
Based on above observations, we propose to use the part composite feature of a 3D shape to recognize unknown sample and its schematic design is illustrated in the orange box in Fig.~\ref{fig: pipeline}. 

Give a point cloud ${x}\in\mathbb{R}^{L\times~3}$ containing $L$ 3D points. Firstly, we extract point-wise features $Z\in\mathbb{R}^{L\times~D}$ from a point cloud with pointnet++~\cite{qi2017pointnet++}, where $D$ is the dimension of point cloud features.
Then, we construct $M$ learnable part prototypes for each known class to enable class-specific part prototypes, which is denoted with $\mathcal{P} = \{p^{k}_{m}\in\mathbb{R}^{D}\}_{k,m=1}^{K,M}$ for $K$ known classes. 
Part prototype $p^{k}_{m}$ encodes some prototypical part information of a 3D shape with class $k$.
As illustrated on the top of Fig.~\ref{fig: pipeline}, part prototypes of a table represent \emph{a table plane, corners,} and \emph{legs}, respectively. 

Then we project point-wise features into the part embedding space according to the similarity between the point cloud features and the part prototype as follows:
\begin{equation}
      Z_p = SZ, S = h(\mathcal{P}Z^T),
    \label{eq:similarity}
\end{equation}
where $S \in \mathbb{R}^{KM\times L}$ is a similarity map between part prototypes and point features and $h(\cdot)$ is a softmax function to produce normalized probabilities along each column. Then we can aggregate local features in $Z$ into part composite features $Z_p\in\mathbb{R}^{KM\times D}$ according to the similarity map $S$.
In order to reduce the computation cost and memory footprints, we feed $Z_p$ to a multi-layer perceptron with channels ($D, D_r$) to reduce the feature dimension, then we concatenate output features and produce part-ware feature $Z_c\in\mathbb{R}^{1 \times D_c}$, where $D_c = KM \times D_r$.


Finally, we tackle 3D open-set recognition following a prototype-based method~\cite{yang2020GCP}. 
We define a set of learnable class-specific prototypes $\mathcal{C}=\{\mathcal{C}^k \in \mathbb{R}^{1 \times D_c}\}_{k=1}^K$ for $K$ known classes. 
The network can be optimized by minimizing the multi-class cross entropy loss $L_{ce}$ and the regularization loss $L_{pl}$.
Loss $L_{ce}$ serving for known class classification is defined as follows:
 \begin{equation}
    \label{eq:loss_ce}
        L_{ce}(Z_c,\mathcal{C}) =-log \frac{e^{-d(Z_c,\mathcal{C}^i})}{\sum^{K}_{k=1}e^{-d(Z_c,\mathcal{C}^k})},
\end{equation}
where the composite feature $Z_c$ belongs to class $i$, $d(Z_c,\mathcal{C}^i)$ is the cosine similarity between $Z_c$ and class-specific prototype $\mathcal{C}^i$.
Regularization loss $L_{pl}$ on closed-set classes makes intra-class features compact to alleviate the overconfidence issue by: 
\begin{equation}
    L_{pl}(Z_c,\mathcal{C}) = ||Z_c-\mathcal{C}^i||_2^2. 
    \label{eq:loss_cons}
\end{equation}

\subsection{Constraining Part Prototypes}\label{subsec:part-constr}
\begin{figure}[t] 
    \includegraphics[width=\linewidth]{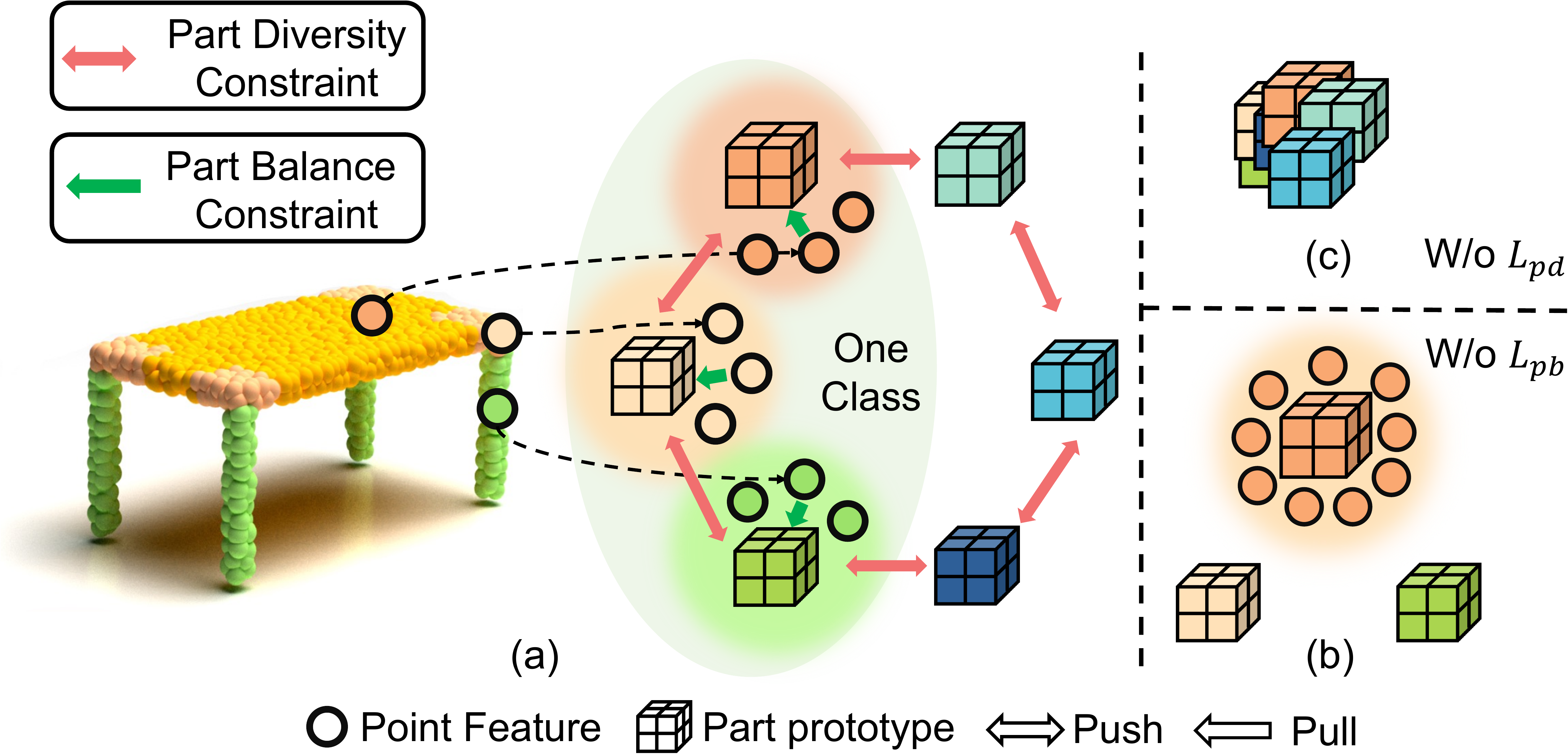}
\caption{(a) Part balance constraint $L_{pb}$ forces the points of a table uniformly assigned to intra-class part prototypes so that part prototypes can learn balanced local information. Part diversity constraint $L_{pd}$ makes all part prototypes differ from each other. (b) Without $L_{pb}$, a single part prototype may represent all points of a table and degrade to a class-specific prototype. (c) Without $L_{pd}$, some part prototypes may collapse into an identical one. 
}
\label{fig:part_loss} 
\end{figure}

To ensure the representation ability of each known class and the inter-class discrimination of part composite features, part prototypes should satisfy the following assumptions: 
\begin{enumerate}[1)]
    \item \textbf{Part Balance.} Part prototypes of a known class should have similar importance. Therefore, learnt information of intra-class part prototypes should be balanced and cover different local areas in a 3D shape to discover different intra-class part patterns.
    This avoids the trivial solution that all point features are represented by a single part prototype and the others are ignored as shown in Fig.~\ref{fig:part_loss}(b). 
    
    \item \textbf{Part Diversity.} All part prototypes should be diverse enough so that they can represent different shape regions and make the inter-class part compositions more distinctive. This avoids the risk that some part prototypes collapse into an identical one as shown in Fig.~\ref{fig:part_loss}(c).
\end{enumerate}
To this end, we propose two constraints for training part prototypes as shown in Fig.~\ref{fig:part_loss}(a).

\textbf{Part Balance Constraint.}  
Since there are no part supervision, we force the number of points in a shape assigned to intra-class part prototypes to be uniform to ensure the part prototypes can learn balanced local information. 
The problem is equivalent to finding an optimal transport from a point cloud to intra-class part prototypes that ensures point features are similar to their corresponding prototypes as much as possible.

Let $Z^k \in \mathbb{R}^{L \times D}$ be learnt point-wise features of a point cloud belonging to class $k$. 
The goal is to find a binary assignment $\Gamma^k = \{\Gamma^k_{i,j} \in [0,1]\}^{L,M}_{i,j=1}$, where $\Gamma^k_{i,j}=1$ means point $i$ is assigned to part prototype $j$. 
Note that one point can only be assigned to a prototype and the number of points assigned to each prototype should be uniform, therefore we have two constraints: $\Gamma^k \mathbf{1}_M = \mathbf{1}_L$ and ${\Gamma^k}^T \mathbf{1}_L= \frac{L}{M}\mathbf{1}_M$.
The optimal assignment should maximize the similarity between each point feature and its assigned prototype, where the similarity is calculated as $Tr({\Gamma^k} {P^k} {Z^k}^T)$. To reduce the time complexity, we use the entropy regularization~\cite{caron2020swav} to obtain an approximate solution.
Therefore, we formulate the problem as follows: 
    \begin{equation}
    \begin{split}
    \label{eq:transport}
     &\max_{\Gamma^k}{Tr({\Gamma^k} {P^k} {Z^k}^T}) + \epsilon H(\Gamma^k),  \\
    s.t.~\Gamma^k &\in [0,1] ^{L\times M}, \Gamma^k \mathbf{1}_M = \mathbf{1}_L, \Gamma^k \mathbf{1}_L= \frac{L}{M}\mathbf{1}_M, 
    \end{split}
\end{equation}
where $H(\Gamma^k)=-\sum_{ij}\Gamma^k_{ij} log \Gamma^k_{ij}$ is the entropy regularization and $\epsilon > 0$ controls the smoothness of the assignment (we set $\epsilon$ to 0.05 empirically).
The problem can be efficiently solved via an iterative Sinkhorn-Knopp algorithm~\cite{cuturi2013sinkhorn} on GPUs. Please see supplementary materials for more details.

After obtaining the optimal assignment matrix $\Gamma^k$, we can enforce the part balance constraint by requiring each point feature $z^k_i \in Z^k$ to be similar to its assigned prototype as much as possible, which is equivalent to minimize the following loss: 
\begin{equation}
    \label{eq:loss_pa}
    L_{pb} = - \frac{1}{K}\sum_{k \in K} \sum_{z^k_i\in Z^k} log\frac{e^{-d(z^k_i,p^{k}_{m_i})/\tau}}{\sum_{p^{k}_{m}\in P^k} e^{-d(z^k_i,p^{k}_{m})/\tau}},
\end{equation}
where $p^{k}_{m_i}$ is the part prototype assigned to $z_i^k$ based on  $\Gamma^k$, $d(\cdot,\cdot)$ measures the cosine similarity between point features and their corresponding prototypes, and $\tau$ is a scalar temperature hyperparameter (we set $\tau$ to 0.1 empirically).

\textbf{Part Diversity Constraint.} 
To further encourage part prototypes to be diverse and to represent different patterns in 3D shapes, we minimize the similarity between different part prototypes with the following loss:
\begin{equation}
    \label{eq:loss_pd}
    L_{pd}=\sum_{k=1}^K \sum_{m=1}^M \max_{p^-\in\{p^k_m\}_{k,m=1}^{K,M}/p^{k}_{m}}\left(0, \frac{{p^{k}_{m}}^T p^{-}}{\left\|p^{k}_{m}\right\|\cdot\left
    \|p^{-}\right\|}-\delta \right),
\end{equation}
where $\delta$ is the cosine similarity threshold between a prototype and the other ones. In our experiments, we set $\delta=0.1$.

\subsection{Part-based Unknown Feature Synthesis}
\label{subsec:part-virtual}

\begin{figure}[t] 
    \includegraphics[width=\linewidth]{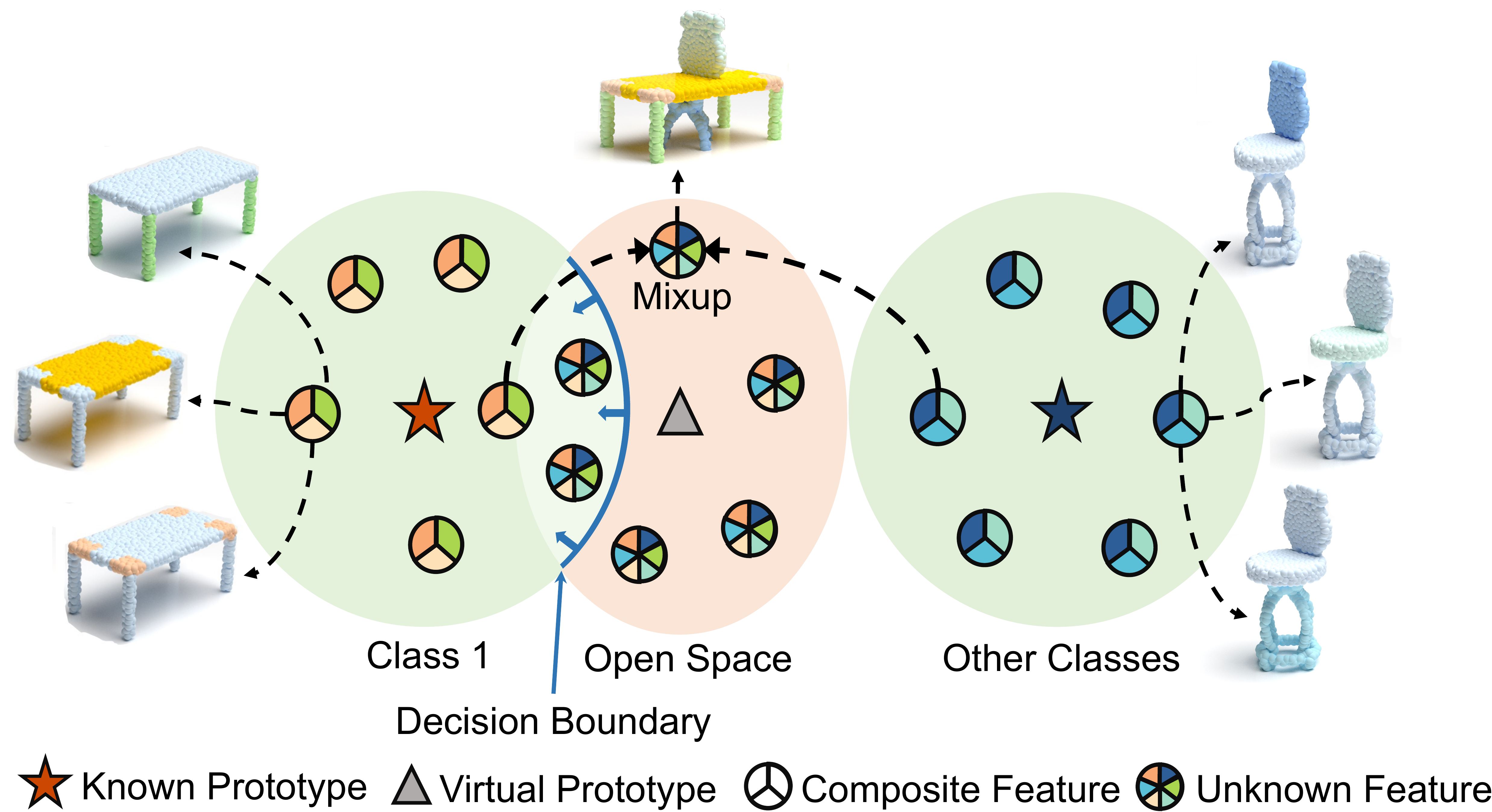}
\caption{An illustration of the part-based unknown feature synthesis. We mix up the part composite features of different known classes to synthesize unknown features near the decision boundary of known classes. Note that synthesized unknown features in the open space consist of parts from different known classes and in most cases represent unknown samples. 
}
\vspace{-5pt}
\label{fig:mixup} 
\end{figure}

Inspired by the idea of synthesizing unknown samples to reduce the open-set risks~\cite{ge2017G-openmax,kong2021opengan}, we propose a \emph{Part-based Unknown Feature Synthesis} module (PUFS) to mix up part composite features of different known classes to synthesize unknown features as the proxy of unknown samples as shown in Fig~\ref{fig:mixup}. The advantages are summarized as follows:

\begin{enumerate}[1)]
    \item \textbf{Efficiency.} Mixing up features is more simple and effective than existing generation-based methods~\cite{ge2017G-openmax}, which require additional generator training.
    \item \textbf{Effectiveness.} Composing shape parts of different classes usually yields samples of novel classes which share some parts with known classes and can enhance unknown sample recognition. 
\end{enumerate}

Formally, giving a part composite feature $Z_{p}^{i}\in\mathbb{R}^{KM\times D}$ of class $i$ in a training batch, our goal is to mix it with another feature $Z_{p}^{j}$ with a different class $j$ to synthesize unknown features:

\begin{equation}
    \label{eq:mix}
    Z_{vp}^{i} = \lambda Z_{p}^{i} + (1-\lambda) Z_{p}^{j},
\end{equation}
where $Z_{vp}^{i} \in \mathbb{R}^{KM\times D}$ denotes a virtual unknown feature share some similar part features with $Z_{p}^{i}$. We sample $\lambda$ from an uniform distribution $[0.6, 1.0]$ to ensure the virtual unknown features are near the decision boundary of category $i$.
Then we can obtain $K$ kind of unknown features similar to each known class. 
Note that synthesizing unknown features following Eq.(\ref{eq:mix}) is efficient and effective. 
Afterwards, we concatenate $Z_{vp}^{i}$ and output part-ware feature $Z_{vc}^{i}$.

To learn the classification of these unknown features, we construct $K$ learnable virtual unknown prototypes $\mathcal{V} = \{\mathcal{V}^{k}\in \mathbb{R}^{1 \times D_c}\}_{k=1}^K$ to represent $K$ kinds of unknown features. Each virtual unknown prototype learns the distribution of unknown features near the decision boundary of a known class.
We extend Eq.(\ref{eq:loss_ce}) with the virtual unknown prototypes $\mathcal{V}$ as follows:
\begin{equation}
    \label{eq:loss_vs}
    \begin{split}
        L_{ce}(Z_{vc}^i;C, V) =- log \frac{e^{-d(Z_{vc}^{i}, \mathcal{V}^{i})}}{\sum^{K}_{k=1}e^{-d(Z_{vc}^{i},\mathcal{C}^k\cup\mathcal{V}^k)}},
    \end{split}
\end{equation}
where $\mathcal{C}^k\cup\mathcal{V}^k$ is the union of known class prototypes and virtual unknown prototypes. 
By optimizing this loss, the classifier can obtain more compact decision boundaries for known classes and manage open risks through virtual unknown prototypes.

\subsection{Network Learning}
In summary, the overall training objective loss function is:
\begin{equation}
    L = L_{ce} + \lambda_1\cdot{}L_{pl} + \lambda_2\cdot{}L_{pb} + \lambda_3\cdot{}L_{pd},
\end{equation}
where the classification loss $L_{ce}$ in Eq.(\ref{eq:loss_vs}) minimizes the risk on closed-set classes, regularization loss $L_{pl}$ in Eq.(\ref{eq:loss_cons}) reduces intra-class feature variances, loss $L_{pb}$ in Eq.(\ref{eq:loss_pa})  and $L_{pd}$ in Eq.(\ref{eq:loss_pd}) enforces basic assumptions of part prototypes. 
The last three term avoids the overconfidence of closed-set classifiers as well as the under-representation problem of class-specific prototype methods.
Weight term $\lambda_*$ balances different losses and are set to ($0.1, 1.0, 1.0$).

During inference, giving a test sample's feature $Z_c^*$, we use the cosine similarity between feature and all prototypes for unknown samples detection. 
\begin{equation}
y^* = \left\{
\begin{aligned}
K + 1 & , & \mathrm{if}~k^* > K, \\
k^*  & , & \mathrm{otherwise},
\end{aligned}
\right.
\end{equation}
where $k^* = \underset{k=1,...,2K}{\mathrm{argmax}} [d(\mathcal{C}\cup\mathcal{V}, Z_c^*)]$ is the predicted label of a sample and $K+1$ indicates the unknown class.


    

\section{Experiments} \label{sec:results}

\begin{table*}[t]
\caption{The ACC and OSCR results on single dataset tasks. The results are averaged among five randomized trials.}
\label{tab:single_dataset}
\resizebox{\linewidth}{!}{%
\begin{tabular}{l|cc|cc|cc|cc|cc|cc}
\toprule
        &\multicolumn{2}{c}{ModelNet 5-35}&  \multicolumn{2}{c}{ModelNet 20-20}  &\multicolumn{2}{c}{ShapeNet 5-50} & \multicolumn{2}{c}{ShapeNet 20-35}  & \multicolumn{2}{c}{ScanObjectnn 5-10} & \multicolumn{2}{c}{ScanObjectNN 10-5}\\ \midrule
        & ACC  & OSCR  & ACC  & OSCR & ACC  & OSCR 
        & ACC  & OSCR  & ACC  & OSCR  & ACC  & OSCR \\ \midrule
Softmax  & 100 & 88.9 & 93.9 & 86.4 & 99.2 & 87.5 & 98.8 & 87.0 & 94.7 & 82.7 & 84.0 & 70.4   \\
GCP~\tiny(PAMI, 2020)     & 100 & 88.4 & 96.4 & 85.0 & 99.1 & 90.7 & 99.1 & 87.1 & 94.4 & 78.3 & 83.9 & 66.3   \\
RPL~\tiny(ECCV, 2020)       & 100 & 88.6 & 95.9 & 86.4 & 98.8 & 88.6 & 97.8 & 86.9 & 93.7 & 80.4 & 83.6 & 66.0   \\
PROSER~\tiny(CVPR, 2021)    & 100 & 89.2 & 95.6 & 87.0 & 98.8 & 88.0 & 99.3 & 88.7 & 93.8 & 80.5 & 84.2 & 72.7   \\
ARPL~\tiny(PAMI, 2021)      & 100 & 91.3 & 95.6 & 89.1 & 99.2 & 88.6 & 98.8 & 93.0 & 95.6 & 84.7 & 85.5 & 75.8   \\ \midrule
Ours     & 100 & \textbf{100 \textcolor{blue}{+ 8.7}}  & \textbf{96.8 \textcolor{blue}{+ 1.2}}  & \textbf{96.7 \textcolor{blue}{+ 7.6}} 
         & \textbf{99.2} & \textbf{99.2 \textcolor{blue}{+ 10.6}} & \textbf{99.5 \textcolor{blue}{+ 0.2}} & \textbf{99.4 \textcolor{blue}{+ 6.4}}
         & \textbf{97.0 \textcolor{blue}{+ 1.4}} & \textbf{97.1 \textcolor{blue}{+ 12.4}} & \textbf{86.8 \textcolor{blue}{+ 1.3}} & \textbf{86.6 \textcolor{blue}{+ 10.8}} 
          \\ 
\bottomrule
\end{tabular}
}
\vspace{-5pt}
\end{table*}

\begin{table*}[t]
\caption{The OSCR results on cross datasets tasks.}
\label{tab:cross_dataset}
\resizebox{\linewidth}{!}{%
\begin{tabular}{l|c|c|c|c|c|c}
\toprule
        &ModelNet $\rightarrow$ ScanObjectNN&   
        ModelNet $\rightarrow$ ShapeNet & 
        ScanObjectNN $\rightarrow$ ShapeNet &
        ScanObjectNN $\rightarrow$ ModelNet &
        ShapeNet $\rightarrow$ ScanObjectNN &
         ShapeNet $\rightarrow$ ModelNet 
        \\ \midrule

Softmax  & 95.3 & 92.1 & 75.2 & 72.6 & 87.2 & 79.7 \\
GCP~\tiny(PAMI, 2020)     & 92.5 & 91.2 & 67.5 & 64.6 & 84.2 & 81.5  \\
RPL~\tiny(ECCV, 2020)       & 92.6 & 93.2 & 71.8 & 68.1 & 84.5 & 83.3  \\
PROSER~\tiny(CVPR, 2021)   & 93.9 & 92.0 & 72.6 & 72.5 & 85.6 & 84.8  \\
ARPL~\tiny(PAMI, 2021)      & 95.6 & 92.4 & 82.3 & 79.8 & 89.1 & 86.0 \\ \midrule
Ours     & \textbf{100 \textcolor{blue}{+ 2.4}} & \textbf{100 \textcolor{blue}{+ 3.8}} & \textbf{91.6 \textcolor{blue}{+ 9.3}} & \textbf{91.6 \textcolor{blue}{+ 11.8}} & \textbf{99.1 \textcolor{blue}{+ 10.0}} & \textbf{99.1 \textcolor{blue}{+ 13.1}}\\ 
\bottomrule
\end{tabular}
}
\vspace{-5pt}
\end{table*}

\begin{table}[t]
\caption{The OSCR results on confusing mixup tasks.}
\label{tab:mix_dataset}
\resizebox{\linewidth}{!}{%
\begin{tabular}{l|c|c|c}
\toprule
        &ModelNetMix&   ScanObjectNNMix & ShapeNetMix\\ \midrule
        
Softmax  & 80.0 & 59.1 & 60.3\\
GCP~\tiny(PAMI, 2020)     & 63.8 & 56.5 & 54.2\\
RPL~\tiny(ECCV, 2020)  & 81.4 & 50.0 & 52.9\\
PROSER~\tiny(CVPR, 2021)   & 76.0 & 58.6 & 56.4\\
ARPL~\tiny(PAMI, 2021)     & 75.6 & 58.7 & 58.4\\ \midrule
Ours     & \textbf{98.0 \textcolor{blue}{+ 16.6}} & \textbf{91.5 \textcolor{blue}{+ 32.4}}& \textbf{99.1 \textcolor{blue}{+ 38.8}}\\ 

\bottomrule
\end{tabular}
}
\vspace{-5pt}
\end{table}

\subsection{Evaluation Datasets for 3D OSR}
We construct three OSR tasks to evaluate the performance of 3D OSR based on existing 3D recognition datasets, \ie, ModelNet40~\cite{wu2015modelnet}, ShapeNetCore~\cite{chang2015shapenet}, and ScanObjectNN~\cite{uy2019scanobjectnn}. 


\textbf{Single Dataset OSR} focuses on evaluating 3D OSR on shapes from the same dataset. 
In this way, the testing data has a similar geometry distribution as the training data but contains unknown shapes. 
We randomly sample $K$ classes to be known and the other $U$ classes to be unknown and denote it with DATASET-K-U. For example, ModelNet-$5$-$35$ means the dataset takes $5$ known classes and $35$ unknown classes in ModelNet.

\textbf{Cross Dataset OSR} assesses the performance of 3D OSR on cross-domain shape recognition. 
This reflects the real-world scenarios that open set samples may come from different domains. 
In this task, we set five common classes of the three datasets, \ie, \emph{bed, chair, sofa, table}, and \emph{bookshelf}, to known and the others to unknown.
The training set collects samples with a known category from a dataset, while the testing set contains samples with a known category from the same dataset and samples with a unknown category from another dataset. 
For example, ModelNet$\rightarrow$ScanObjectNN denotes a dataset with the training data from known classes of ModelNet and the unknown testing data from ScanObjectNN.

\textbf{Confusing Mixup OSR} tests if a 3D OSR method can identify open set samples that share some semantically meaningful parts with known classes~\cite{chen2021ARPL}. 
This kind of open-set samples are common in the real world. For example, both a laptop and a monitor have a screen. 
Directly collecting such a dataset is tedious and we instead synthesize such a dataset by rigid subset mix~\cite{lee2021rsmix} of two samples from five common known classes.

\subsection{Evaluation Metrics}
Following 2D OSR tasks~\cite{dhamija2018reducing}, we take a threshold-independent metric, \ie, Open set Classification Rate (\textbf{OSCR}), to evaluate the performance on identifying unknown samples. 
The OSCR value is calculated by taking the area under the curve of the Correct Classification Rate (CCR) against the False Positive Rate (FPR). We give detailed formulation of this metric in the supplementary. 
In general, a larger value of the OSCR indicates better performance on 3D OSR. 
We also evaluate the performance of known classes by calculating the classification accuracy (\textbf{ACC}) 

\subsection{Comparison with SOTA methods}
As far as we know, this work makes a first attempt to 3D OSR and there is no previous work on this problem. 
Existing works on 2D OSR provide many inspirations for its 3D counterparts. In this section, we adapt 2D OSR methods to 3D recognition and select two kinds of SOTA methods as the baselines, \ie, prototype-based methods (GCP~\cite{yang2020GCP}, RPL~\cite{chen2020RPL}, and ARPL~\cite{chen2021ARPL}) and the calibration-based PROSER~\cite{zhou2021placeholder}.  
Basically, we replace the image-based backbone of these methods with a 3D one (\ie, Pointnet++ in our experiments) and employ the same pipeline for 3D OSR.  
We also compare with a closed-set classifier denoting with \emph{softmax}.

\textbf{Results on the Single Dataset OSR} is shown in Tab.~\ref{tab:single_dataset}. 
For the closed-set recognition, all methods perform quite well and can achieve an accuracy greater than $90\%$ on both ModelNet and ShapeNet. 
Nevertheless, our method is still the winner on all configurations. 
For the open-set recognition, we can see PROSER and APRL achieve better performance than the other baselines in most configurations. 
Our method consistently outperforms all baselines by a large margin on all kinds of configurations. 
This indicates the overwhelming advantage of our method over the others on 3D OSR of samples from a similar distribution. 


\textbf{Results on Cross Dataset OSR} are presented in Tab.~\ref{tab:cross_dataset}. 
Note that we only show the performance of OSCR because samples in the testing dataset is the same as single dataset OSR and their results are shown in Tab.~\ref{tab:single_dataset}. 
We can observe that compared methods lack robustness on different kinds of datasets and their performance vary a lot depending on the dataset configuration. 
Our method performs consistently well among different datasets and outperforms existing methods by a large margin ($8.4\%$ on average).
We believe the good performance is owing to part prototypes, which can generalize to different kinds of data. 

\textbf{Results on Confusing Mixup OSR} are given in Tab.~\ref{tab:mix_dataset}. 
It can be observed that prototype-based methods suffer from severe performance degradation. 
This exposes the under-representation of prototype-based methods that only focus on partial regions and filter out necessary shape parts that can help distinguish unknown shapes as shown in Fig.~\ref{fig1}.
The calibration-based method, \ie, PROSER, also has difficulty in recognizing samples with similar parts.  
Our method can overcome the under-representation by introducing part prototypes, which yields excellent results and improves the performance a lot ($29.3\%$ on average).

\subsection{Ablation Study}
In this section, we evaluate the influence of different modules on the ModelNet-$20$-$20$ dataset. We adopt GCP~\cite{yang2020GCP} as baseline and gradually adding our proposed modules to build our method.

\textbf{Ablation on Model Components.}
Our method consists of several important parts, including part prototype-based classifier, a part prototype diversity constraint loss $L_{pd}$, a part prototype balance constraint loss $L_{pb}$, concatenation for feature fusion, and part-based unknown feature synthesis (PUFS). As shown in Tab.~\ref{tab:ablation}, we validate the impact of each part by adding the corresponding part step by step or replacing a module with another design.

We can see that without two constraints on part prototypes, adding part prototypes only achieves very small improvement comparing to the baseline. Adding $L_{pd}$ and $L_{pb}$ results in $9.2$ points gain on OSCR. This indicates that part prototypes with proper constraints can encode more information of a shape and alleviate the under-representation problem in previous prototype-based methods.
If we replace the concatenation operation with a max pooling operation, the performance suffers from a drop ($1.4$ points). 
This means that the concatenation of features is a better way to preserve part information. 
The PUFS can also increase the performance by about $0.9$ points on OSCR, which implies the effectiveness of synthesized unknown features. 

    \begin{table}[t]

\caption{Ablation Experiments}
\label{tab:ablation}
\resizebox{\linewidth}{!}{%
\begin{tabular}{ccccc|c}
\toprule  part prototype & $L_{pd}$ & $L_{pb}$ & cat$\rightarrow$ max  & PUFS & OSCR  \\ \midrule
           &           &           &           &           & 85.5    \\     
\checkmark &           &           &           &           & 86.6    \\
\checkmark &\checkmark &           &           &           & 93.9    \\
\checkmark &           &\checkmark &           &           & 93.6    \\
\checkmark &\checkmark &\checkmark &           &           & 95.8    \\
\checkmark &\checkmark &\checkmark &\checkmark &           & 94.4    \\
\checkmark &\checkmark &\checkmark &           &\checkmark & 96.7    \\  
\bottomrule
\end{tabular}
}
\vspace{-10pt}
\end{table}

\begin{figure}[t] 
    \centering
    \subfloat[Number of Part Prototypes]{\label{fig:part_num} 
    \includegraphics[width=0.5\linewidth]{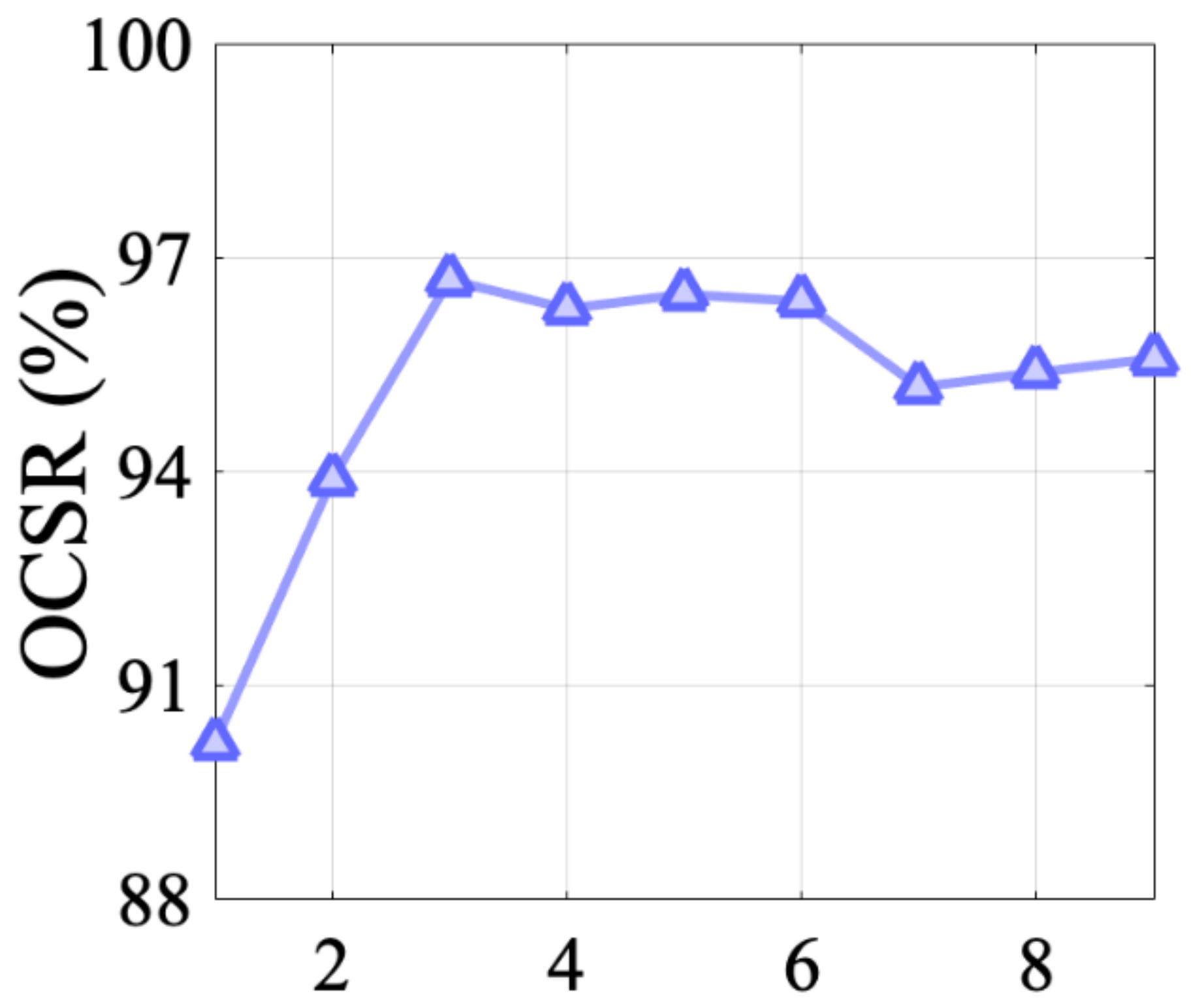}
    }
    \subfloat[Cosine Similarity $\delta$]{
    \label{fig:cos_sim}
    \includegraphics[width=0.5\linewidth]{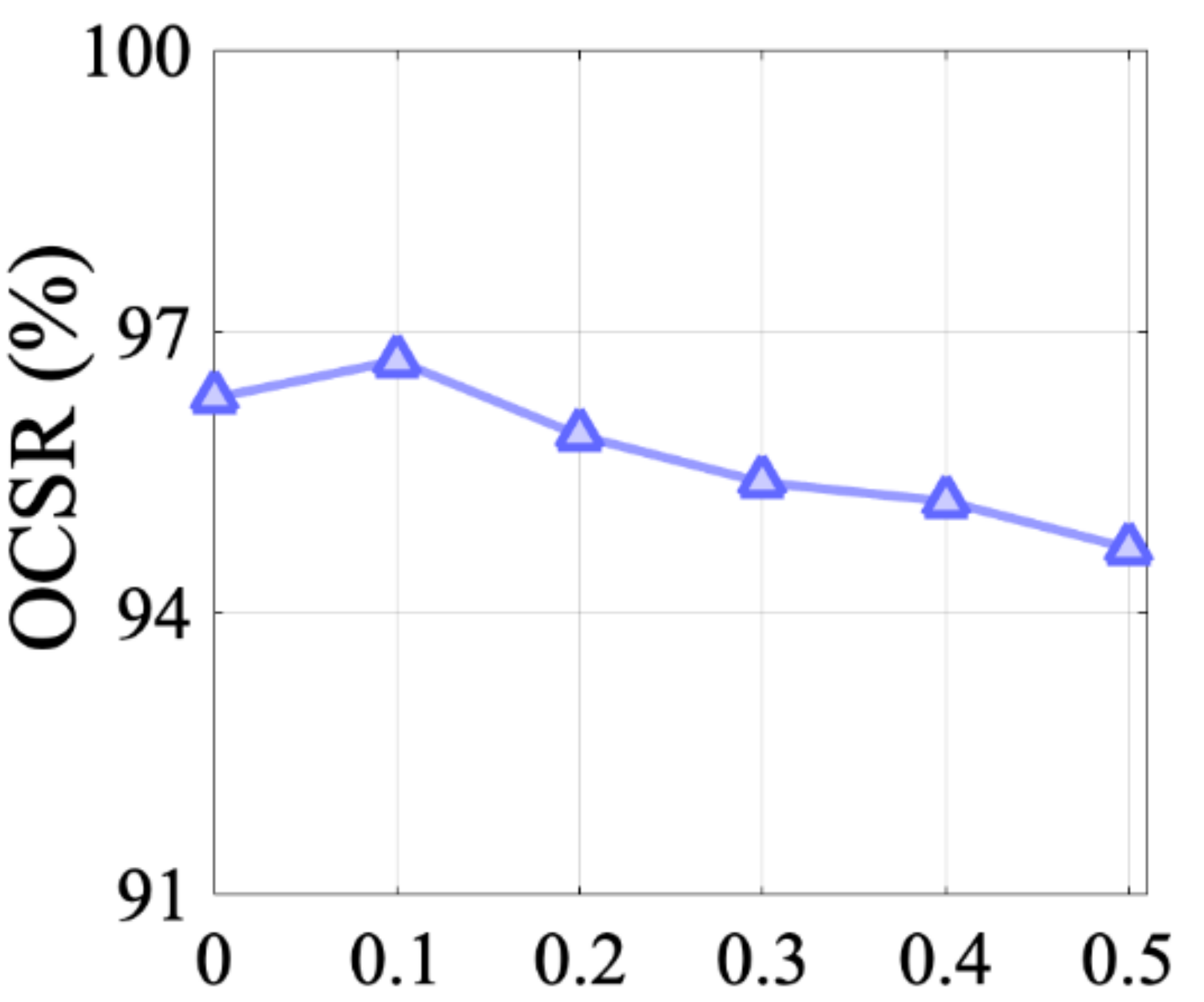}
    }
\caption{Evaluation of hyper-parameters. We show results on (a) different number of part prototypes for each class and (b) the cosine similarity threshold $\delta$ in Eq.\ref{eq:loss_pd}}
\label{fig:param} 
\vspace{-5pt}
\end{figure}

\textbf{Number of part prototypes.} We evaluate the number of part prototypes $M$ for each class in Fig.~\ref{fig:part_num}. We notice that when $M=1$, the result is similar to baseline, and when $M\ge3$, the results start to saturate. 

\textbf{Cosine similarity threshold.} We assess the cosine similarity threshold $\delta$ in Eq.\ref{eq:loss_pd}. As shown in Fig.~\ref{fig:cos_sim}, our method achieves the best performance when $\delta=0.1$. When increasing the threshold $\delta$, the part prototypes tend to be identical and the results decrease continuously.

\begin{figure}[H] 
    \centering
    \includegraphics[width=\linewidth]{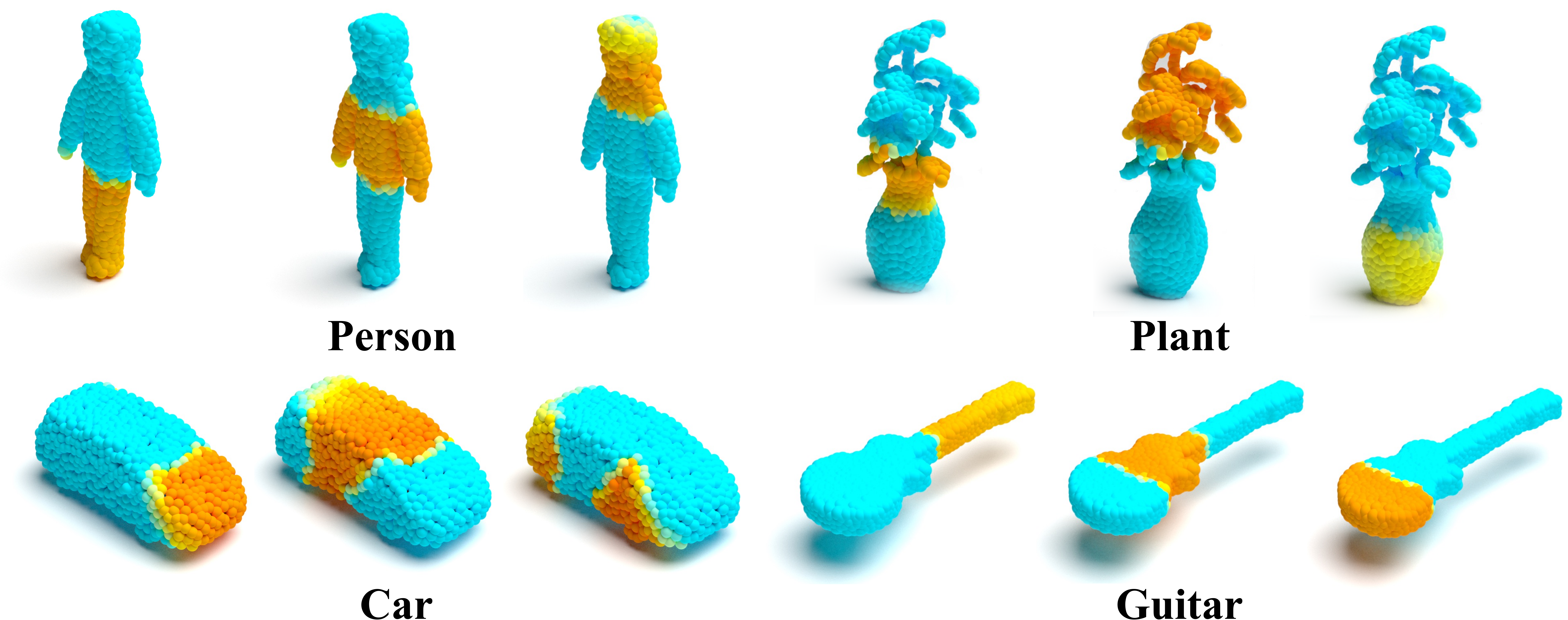}
\caption{Visualization of the similarity map $S$.}
\label{fig:attention map} 
\end{figure}

\subsection{Similarity Map Visualization.}
As shown in Fig.~\ref{fig:attention map}, we visualize the similarity map $S$ in Eq.\ref{eq:similarity} between point cloud features and their corresponding part prototypes. 
We can see different part prototypes can focus on different shape regions. 
Also the composition of shape parts depends on the shape category. 
We show more visualizations in the supplementary.



        

\section{Conclusion}\label{sec:future}
This paper investigates 3D open-set recognition. We propose a part prototype-based 3D OSR method that uses part composite features to solve the overconfidence and under-representation problem of existing methods. Then we develop two constraints to ensure the effectiveness of part prototypes and a PUFS module to synthesize unknown samples efficiently.
We construct various 3D OSR tasks for evaluation. Extensive experiments show that the method overwhelms all baselines consistently on all 3D OSR tasks. 
We hope our work can inspire more researches on 3D OSR and reduce the serious misfortune caused by open-set risks. 

{\small
\bibliographystyle{ieee_fullname}
\bibliography{srcs/openset, srcs/part_learning, srcs/pointcloud}

\begin{thebibliography}{10}\itemsep=-1pt

\bibitem{bendale2016towards}
Abhijit Bendale and Terrance~E Boult.
\newblock Towards open set deep networks.
\newblock In {\em Proceedings of the IEEE conference on computer vision and
  pattern recognition}, pages 1563--1572, 2016.

\bibitem{caron2020swav}
Mathilde Caron, Ishan Misra, Julien Mairal, Priya Goyal, Piotr Bojanowski, and
  Armand Joulin.
\newblock Unsupervised learning of visual features by contrasting cluster
  assignments.
\newblock {\em Advances in Neural Information Processing Systems},
  33:9912--9924, 2020.

\bibitem{chang2015shapenet}
Angel~X Chang, Thomas Funkhouser, Leonidas Guibas, Pat Hanrahan, Qixing Huang,
  Zimo Li, Silvio Savarese, Manolis Savva, Shuran Song, Hao Su, et~al.
\newblock Shapenet: An information-rich 3d model repository.
\newblock {\em arXiv preprint arXiv:1512.03012}, 2015.

\bibitem{chen2021ARPL}
Guangyao Chen, Peixi Peng, Xiangqian Wang, and Yonghong Tian.
\newblock Adversarial reciprocal points learning for open set recognition.
\newblock {\em arXiv preprint arXiv:2103.00953}, 2021.

\bibitem{chen2020RPL}
Guangyao Chen, Limeng Qiao, Yemin Shi, Peixi Peng, Jia Li, Tiejun Huang,
  Shiliang Pu, and Yonghong Tian.
\newblock Learning open set network with discriminative reciprocal points.
\newblock In {\em European Conference on Computer Vision}, pages 507--522.
  Springer, 2020.

\bibitem{choy20194minkowski}
Christopher Choy, JunYoung Gwak, and Silvio Savarese.
\newblock 4d spatio-temporal convnets: Minkowski convolutional neural networks.
\newblock In {\em Proceedings of the IEEE/CVF Conference on Computer Vision and
  Pattern Recognition}, pages 3075--3084, 2019.

\bibitem{cuturi2013sinkhorn}
Marco Cuturi.
\newblock Sinkhorn distances: Lightspeed computation of optimal transport.
\newblock {\em Advances in neural information processing systems}, 26, 2013.

\bibitem{dhamija2018reducing}
Akshay~Raj Dhamija, Manuel G{\"u}nther, and Terrance Boult.
\newblock Reducing network agnostophobia.
\newblock {\em Advances in Neural Information Processing Systems}, 31, 2018.

\bibitem{ge2017G-openmax}
ZongYuan Ge, Sergey Demyanov, Zetao Chen, and Rahil Garnavi.
\newblock Generative openmax for multi-class open set classification.
\newblock {\em arXiv preprint arXiv:1707.07418}, 2017.

\bibitem{hermann2020shapes}
Katherine Hermann and Andrew Lampinen.
\newblock What shapes feature representations? exploring datasets,
  architectures, and training.
\newblock {\em Advances in Neural Information Processing Systems},
  33:9995--10006, 2020.

\bibitem{kong2021opengan}
Shu Kong and Deva Ramanan.
\newblock Opengan: Open-set recognition via open data generation.
\newblock In {\em Proceedings of the IEEE/CVF International Conference on
  Computer Vision}, pages 813--822, 2021.

\bibitem{lee2021rsmix}
Dogyoon Lee, Jaeha Lee, Junhyeop Lee, Hyeongmin Lee, Minhyeok Lee, Sungmin Woo,
  and Sangyoun Lee.
\newblock Regularization strategy for point cloud via rigidly mixed sample.
\newblock In {\em Proceedings of the IEEE/CVF Conference on Computer Vision and
  Pattern Recognition}, pages 15900--15909, 2021.

\bibitem{li2018pointcnn}
Yangyan Li, Rui Bu, Mingchao Sun, Wei Wu, Xinhan Di, and Baoquan Chen.
\newblock Pointcnn: Convolution on x-transformed points.
\newblock {\em Advances in neural information processing systems}, 31, 2018.

\bibitem{neal2018counterfactual}
Lawrence Neal, Matthew Olson, Xiaoli Fern, Weng-Keen Wong, and Fuxin Li.
\newblock Open set learning with counterfactual images.
\newblock In {\em Proceedings of the European Conference on Computer Vision
  (ECCV)}, pages 613--628, 2018.

\bibitem{nguyen2015overconf}
Anh Nguyen, Jason Yosinski, and Jeff Clune.
\newblock Deep neural networks are easily fooled: High confidence predictions
  for unrecognizable images.
\newblock In {\em Proceedings of the IEEE conference on computer vision and
  pattern recognition}, pages 427--436, 2015.

\bibitem{qi2017pointnet}
Charles~R Qi, Hao Su, Kaichun Mo, and Leonidas~J Guibas.
\newblock Pointnet: Deep learning on point sets for 3d classification and
  segmentation.
\newblock In {\em Proceedings of the IEEE conference on computer vision and
  pattern recognition}, pages 652--660, 2017.

\bibitem{qi2017pointnet++}
Charles~Ruizhongtai Qi, Li Yi, Hao Su, and Leonidas~J Guibas.
\newblock Pointnet++: Deep hierarchical feature learning on point sets in a
  metric space.
\newblock {\em Advances in neural information processing systems}, 30, 2017.

\bibitem{selvaraju2017grad}
Ramprasaath~R Selvaraju, Michael Cogswell, Abhishek Das, Ramakrishna Vedantam,
  Devi Parikh, and Dhruv Batra.
\newblock Grad-cam: Visual explanations from deep networks via gradient-based
  localization.
\newblock In {\em Proceedings of the IEEE international conference on computer
  vision}, pages 618--626, 2017.

\bibitem{uy2019scanobjectnn}
Mikaela~Angelina Uy, Quang-Hieu Pham, Binh-Son Hua, Thanh Nguyen, and Sai-Kit
  Yeung.
\newblock Revisiting point cloud classification: A new benchmark dataset and
  classification model on real-world data.
\newblock In {\em Proceedings of the IEEE/CVF international conference on
  computer vision}, pages 1588--1597, 2019.

\bibitem{wang2017ocnn}
Peng-Shuai Wang, Yang Liu, Yu-Xiao Guo, Chun-Yu Sun, and Xin Tong.
\newblock O-cnn: Octree-based convolutional neural networks for 3d shape
  analysis.
\newblock {\em ACM Transactions On Graphics (TOG)}, 36(4):1--11, 2017.

\bibitem{wang2019dgcnn}
Yue Wang, Yongbin Sun, Ziwei Liu, Sanjay~E Sarma, Michael~M Bronstein, and
  Justin~M Solomon.
\newblock Dynamic graph cnn for learning on point clouds.
\newblock {\em Acm Transactions On Graphics (tog)}, 38(5):1--12, 2019.

\bibitem{wu2015modelnet}
Zhirong Wu, Shuran Song, Aditya Khosla, Fisher Yu, Linguang Zhang, Xiaoou Tang,
  and Jianxiong Xiao.
\newblock 3d shapenets: A deep representation for volumetric shapes.
\newblock In {\em Proceedings of the IEEE conference on computer vision and
  pattern recognition}, pages 1912--1920, 2015.

\bibitem{yang2020GCP}
Hong-Ming Yang, Xu-Yao Zhang, Fei Yin, Qing Yang, and Cheng-Lin Liu.
\newblock Convolutional prototype network for open set recognition.
\newblock {\em IEEE Transactions on Pattern Analysis and Machine Intelligence},
  2020.

\bibitem{zhao2021pointtransformer}
Hengshuang Zhao, Li Jiang, Jiaya Jia, Philip~HS Torr, and Vladlen Koltun.
\newblock Point transformer.
\newblock In {\em Proceedings of the IEEE/CVF International Conference on
  Computer Vision}, pages 16259--16268, 2021.

\bibitem{zhou2021placeholder}
Da-Wei Zhou, Han-Jia Ye, and De-Chuan Zhan.
\newblock Learning placeholders for open-set recognition.
\newblock In {\em Proceedings of the IEEE/CVF Conference on Computer Vision and
  Pattern Recognition}, pages 4401--4410, 2021.

\end{thebibliography}
}

\end{document}